\documentclass[journal]{IEEEtran}
\usepackage{amsmath,amsfonts}
\usepackage{algorithmic}
\usepackage{algorithm}
\usepackage{array}
\usepackage[caption=false,font=normalsize,labelfont=sf,textfont=sf]{subfig}
\usepackage{textcomp}
\usepackage{stfloats}
\usepackage{url}
\usepackage{color}
\usepackage{verbatim}
\usepackage{graphicx}
\usepackage{multirow}
\usepackage[table,xcdraw]{xcolor}
\usepackage[colorlinks,bookmarksopen,bookmarksnumbered,citecolor=black, linkcolor=black, urlcolor=black]{hyperref}
\usepackage{cite}
\begin{document}

\title{Structured Click Control in Transformer-based Interactive Segmentation}

\author{\IEEEauthorblockN{Long Xu\textsuperscript{1}},
\IEEEauthorblockN{Yongquan Chen\textsuperscript{1}\IEEEauthorrefmark{1}},
\IEEEauthorblockN{Rui Huang\textsuperscript{1}},
\IEEEauthorblockN{Feng Wu\textsuperscript{1}},
\IEEEauthorblockN{Shiwu Lai\textsuperscript{3}}
\thanks{
\noindent\textsuperscript{1} Shenzhen Institute of Artificial Intelligence and Robotics for Society, The Chinese University of Hong Kong, Shenzhen, China.

\textsuperscript{2} University of Science and Technology of China, Hefei, China.

\textsuperscript{3} Maxvision Technology Corp., Shenzhen, China.

\IEEEauthorrefmark{1} Corresponding author: Yongquan Chen (e-mail: yqchen@cuhk.edu.cn).
}
\thanks{This paper was produced by the IEEE Publication Technology Group. They are in Piscataway, NJ.}
\thanks{Manuscript received April 19, 2021; revised August 16, 2021.}}

\markboth{Journal of \LaTeX\ Class Files,~Vol.~14, No.~8, August~2021}%
{Shell \MakeLowercase{\textit{et al.}}: A Sample Article Using IEEEtran.cls for IEEE Journals}

\IEEEpubid{0000--0000/00\$00.00~\copyright~2021 IEEE}

\maketitle

\begin{abstract}
Click-point-based interactive segmentation has received widespread attention due to its efficiency.
However, it's hard for existing algorithms to obtain precise and robust responses after multiple clicks. In this case, the segmentation results tend to have little change or are even worse than before.
To improve the robustness of the response, we propose a structured click intent model based on graph neural networks, which adaptively obtains graph nodes via the global similarity of user-clicked Transformer tokens.
Then the graph nodes will be aggregated to obtain structured interaction features.
Finally, the dual cross-attention will be used to inject structured interaction features into vision Transformer features, thereby enhancing the control of clicks over segmentation results.
Extensive experiments demonstrated the proposed algorithm can serve as a general structure in improving Transformer-based interactive segmentation performance. 
The code and data will be released at \href{https://github.com/hahamyt/scc}{SCC}.
\end{abstract}


\section{Introduction}

Interactive segmentation algorithms based on click points have garnered considerable attention in the field of computer vision for their efficiency.
These algorithms allow users to interact directly with the image data, guiding the segmentation process through a series of clicks. 

However, previous work mostly focused on improving segmentation accuracy with fewer clicks, often neglecting the user's ability to control network output through clicks.
Specifically, this type of method \cite{7780416, 8953578, chen2022focalclick, liu2022simpleclick, ding2022rethinking} typically constructs the user's positive and negative clicks into a simple 2-channel click map, guiding the network to understand interactive information by concatenating it with the input.
Instead, recent works \cite{kirillov2023segment,sam_hq} have integrated click information during decoding, avoiding duplicate encoding to input image.

\begin{figure}[!t]
\centering
\includegraphics[scale=0.58]{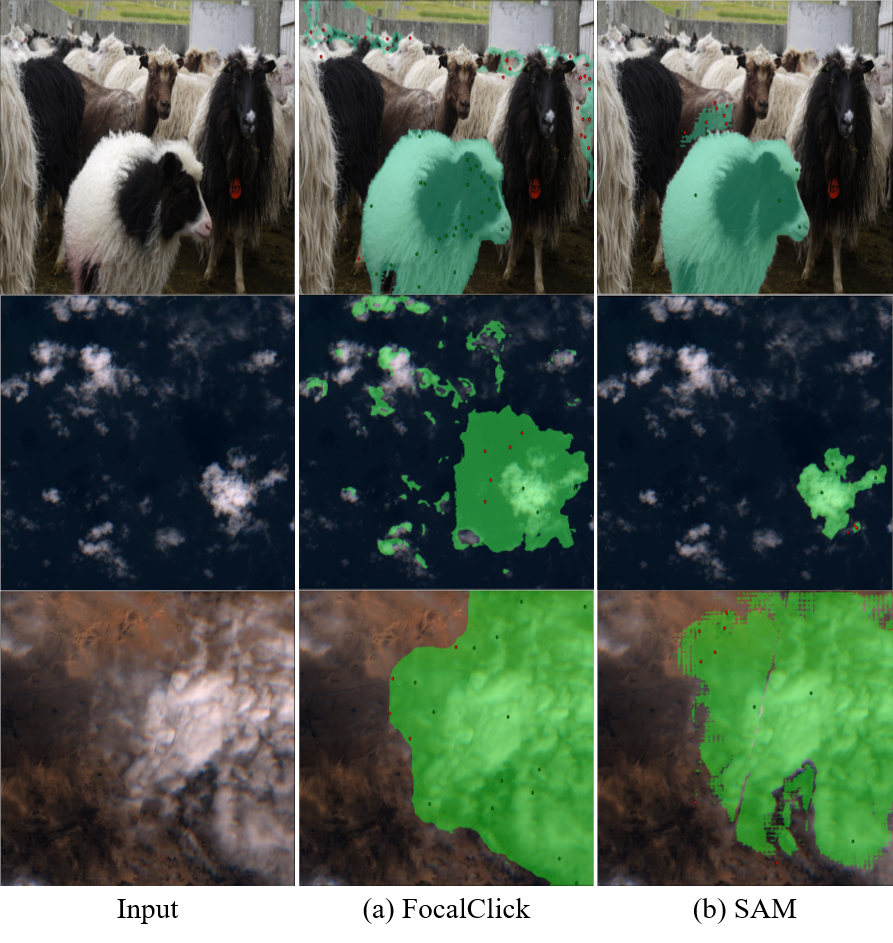}
\caption{The performance of various algorithms after multiple clicks in complex scenarios. The red dots represent negative clicks, while the green dots represent positive clicks. (a) represents a class of click-point modeling algorithms represented by FocalClick \cite{chen2022focalclick}, and (b) represents a class of click-point modeling algorithms represented by SAM. The results indicate that both types of algorithms exhibit issues with uncontrollable network outputs when applied to conventional image data as well as remote sensing data}
\label{fig:motivation}
\end{figure}

Due to the sparse nature of click information, these methods have difficulty understanding user intent and producing robust outputs that meet user expectations.
Furthermore, this can result in the network becoming fixated on local clicked regions, while the output tends to show minimal or even worsening changes compared to the initial state. As shown in Figure \ref{fig:motivation}.

To achieve precise control of network output for user interaction, many solutions have been proposed.

Lin, et al. \cite{9157109} proposed a first click attention algorithm, which improves results with focus invariance, location guidance, and error-tolerant ability. 
However, this algorithm focuses too much on the first click and cannot build user intent under continuous clicks.
Yang, et al. \cite{Yang2023DRE-Net} introduced discriminability in the modeling of user intent, effectively reducing the number of clicks.
Minghao Zhou, et al. \cite{Zhou2023Interactive} proposed a binary classification model based on pixel-level Gaussian processes, which can effectively propagate click information throughout the entire image, thereby enhancing the control of clicks on segmentation results.
Such methods improve the robustness of the network to click categories, but lack effective modeling of user intent.

Ding, et al. \cite{Ding2023Rethinking} proposed the feature-interactive map and interactive nonlocal block, which significantly improved the model's perception of clicks.
Lin, et al. \cite{Lin2021Interactive} proposed a dynamic click transformer network, which improved the performance by employing better click modeling.
Such methods enhance their control over the output by introducing new click models, but lack effective structural information, resulting in insufficient robustness.

To address the problem of insufficient control over model output in existing methods, this paper proposes a structured click control model that uses graph neural networks to learn structural information and utilizes dual cross-attention to fuse structural information with raw features to improve click-to-output control.

The main contributions of this article are as follows:
\begin{itemize}
    \item Proposed a generic click control enhancement algorithm based on GNN, which effectively improves the user's control over network output.
    \item Proposed a dual cross-attention module for effective integration of structured interaction features.
\end{itemize}

\begin{figure*}[!ht]
\centering
\includegraphics[scale=0.78]{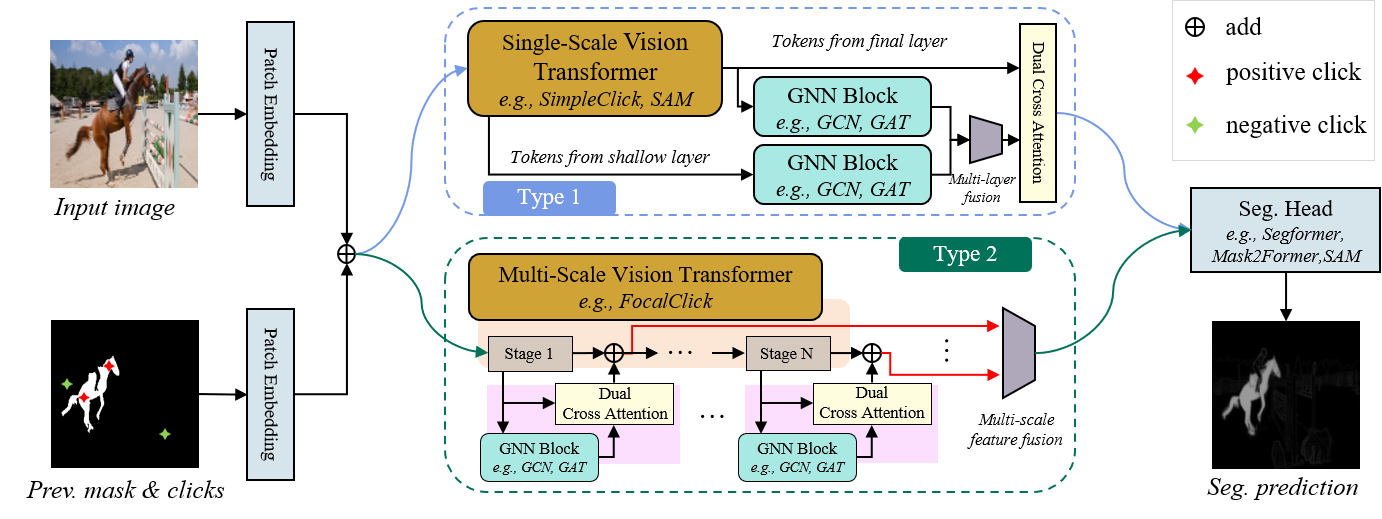}
\caption{Overall structure of the proposed algorithm. (1). Input of the algorithm is including input image, previous mask, and click maps. The patch embedding module will encoding them as tokens then input to the ViT backbone.
(2). 
The most common used ViT are divided into two types including 'Single-Scale' and 'Multi-Scale', which can adopt the proposed algorithm easily. }
\label{fig:framework}
\end{figure*}

\section{Related Work}
The proposed algorithm is based on the patchfied structure of Vision Transformer (ViT), and therefore it is not applicable to CNN-based algorithms.
Therefore, related work includes Transformer-based interactive segmentation and graph neural networks for structuring and modeling interactive information.

\subsection{Interactive segmentation based on transformer}

SimpleClick \cite{liu2022simpleclick} is an interactive image segmentation algorithm with a ViT backbone.
To obtain multi-scale features, SimpleClick leverages SimpleFPN \cite{li2022exploring} to transform image features into four scales during the decoding stage. Additionally, it utilizes a segmentation head similar to Segformer \cite{xie2021segformer} to obtain the final segmentation result.
Focalclick \cite{chen2022focalclick} utilizes Segformer as its primary segmentation framework. Furthermore, it incorporates a compact refiner network to safeguard the stability of the segmented results. 
Both methods use 2-channel click maps to model user interactions, based on these, many related works have been derived \cite{lin2023adaptiveclick, sun2023cfr}.

SAM \cite{kirillov2023segment} and HQ-SAM \cite{sam_hq} both use ViT as image encoder. 
Unlike SimpleClick, which processes click information through the image encoder, the click information in this approach is handled directly in the decoder. While this method may offer higher efficiency when multiple clicks are made, it does not guarantee the stability of the segmentation results.
At the same time, its input size is fixed at $1024\times 1024$, which results in a decrease in efficiency compared to other algorithms.

\subsection{Graph Neural Network}
Through an analysis of the structure of the ViT, we discovered that different tokens can be viewed as graph nodes, which are fused by self-attention for downstream tasks.
However, self-attention relies heavily on the position of tokens, and it is difficult to incorporate the priori of importance based on user clicks.
It is difficult to construct structured interactive features.

An intuitive solution is to use Graph Convolutional Network (GCN) \cite{kipf2016semi} to fuse interactive feature nodes.
However, GCN relies heavily on the structure of the graph \cite{velivckovic2017graph}, and the graph nodes of interaction features are dynamically selected, which cannot guarantee the consistency of the graph structure.

Graph Attention Networks (GATs) proposed by Veli{\v{c}}kovi{\'c} Petar, et al. \cite{velivckovic2017graph} is a better choice.
Unlike GCN, which relies on the Laplacian matrix and degree matrix under a fixed graph structure, GAT is structure-independent when aggregating node features, requiring only consideration of the features of neighboring nodes.

In addition, Yun Seongjun et al. \cite{yun2022FastGTN} proposed Graph Transformer Networks (GTN), which can learn effective node representations on graphs by identifying useful meta-paths. However, GTN requires prior edge information, which is not applicable to our dynamic structured graph.

\section{Method}
The overall framework of the proposed algorithm is shown in Figure \ref{fig:framework}.

In Transformer-based interactive segmentation, input image $x_{img}\in\mathbb{R}^{B\times W\times H}$ will be divided into $L={WH}/{s^2}$ patches according to patch size $s$.
These patches will be encoded by the patch embedding module into tokens, which will be processed by single (multi-)scale self-attention to obtain vision transformer feature $x_f\in\mathbb{R}^{B\times L\times C}$.
$C$ denotes the dimension of the token.

Among the tokens corresponding to these patches, we believe that those that are clicked are more important.
Based on this assumption, we construct a general interactive segmentation framework based on graph neural networks.

\begin{figure}[!h]
\centering
\includegraphics[scale=0.55]{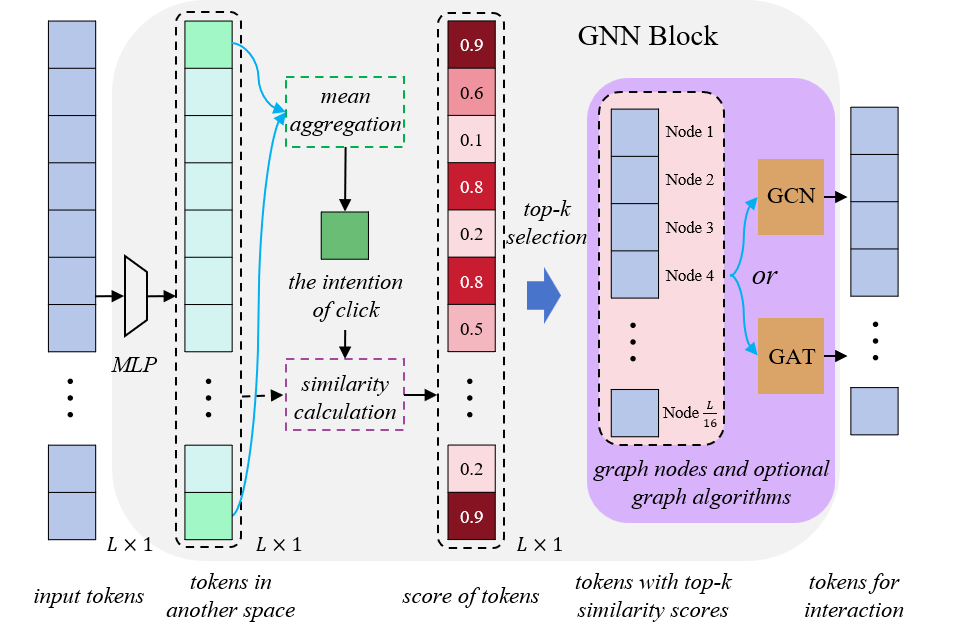}
\caption{Overall structure of the GNN Block. "Top-k selection" refers to selecting tokens with scores greater than 0.95, and the number of tokens selected does not exceed 32. In situations where tokens have fewer similarities with the clicked token, the number of selected tokens is also significantly less than 32. In the GNN Block, any graph neural network can be chosen for node feature fusion. In this paper, we select GCN and GAT.}
\label{fig:gnnblock}
\end{figure}

\subsection{ Graph nodes selection based on user click intention }

We represent the clicked tokens as $x_{c}^{l}\in\mathbb{R}^{B\times \Omega^l\times C}$, where $l\in[0,1]$ represents the click category (positive or negative), 
$\Omega^l$ represents the set of $B$ different clicks under the current category.

Based on $x_c^l$, we can get the average feature $x_a^l\in\mathbb{R}^{B\times C}$ of positive and negative clicks respectively. These features are used to represent the user's click intention, as illustrate in formula (\ref{eq:meanclick}).

\begin{equation}
    \label{eq:meanclick}
    x_a^{l,b}=\frac{1}{|\Omega^l|}\sum_{i\in\Omega^l}x_c[b,i,:],\quad b\in [1,2,...,B]
\end{equation}
\noindent where $x_a^{l,b} $ represents the $b$th click intent feature of class $l$. This intent feature will be used to filter out tokens that might be of interest to the user, as shown in formula (\ref{eq:simselect}).

\begin{equation}
    \label{eq:simselect}
    s^l=\text{sigmoid}(\omega(\tilde{x}_c^l, \tilde{x}_a^{l,b}))
\end{equation}
\noindent where $s^l\in\mathbb{R}^{B\times N}$ denotes the score of all tokens associated with the $l$th class of click intent, which is used to select tokens that match the hypothesis as graph nodes, and $N$ denotes the number of tokens.
$\tilde{x}_c^l=\phi(x_c^l)\in\mathbb{R}^{B\times\Omega^l\times 2C}$, $\tilde{x}_a^{l,b}=\phi({x}_a^{l,b})\in\mathbb{R}^{B\times1\times 2C}$.
$\phi(\cdot)$ represents a simple two-layer feedforward neural network (FNN), which is used to project similarity calculation to another space.
$\omega(\cdot,\cdot)$ is a one-layer FNN for importance score calculation.

The formula (\ref{eq:simloss}) allows the network to learn a robust similarity computation capability.
\begin{equation}
    \label{eq:simloss}
    \mathcal{L}_s=\frac{1}{2}\sum_{l=0}^{1}\frac{\|s^l-y^l\|_2^2}{N}
\end{equation}
\noindent where $y^l$ denotes the importance score label for different categories and $N$ denotes the number of tokens.

\subsection{Incorporating GNN-Derived Structures into Token Interactions}

We use Pytorch's top-k function to get the highest $L/16$ scores $s_k^l\in\mathbb{R}^{B\times \frac{L}{16}}$ and corresponding tokens.
These tokens serve as a collection of graph nodes $\mathcal{V}^{l}\in\mathbb{R}^{B\times\frac{L}{16}\times C}$ to obtain interactive features.

During training, due to the heterogeneity of different images, we use parallelization (vmap) to process batch data.
For convenience of description, we disregard the batch size in the following, and use $V^l\in\mathbb{R}^{\frac{L}{16}\times C}$ and $A^l\in\mathbb{R}^{\frac{L}{16}\times \frac{L}{16}}$ to denote the node and neighbour matrix of a sample, respectively.

Due to the lack of an adjacency matrix, graph neural networks cannot be directly used for feature fusion.

Therefore, we propose a node similarity assumption: 
\begin{itemize}
    \item The higher the similarity between two nodes, the closer their relationship is, and the greater the weight of the corresponding edge. 
    \item The lower the similarity, the smaller the weight of the corresponding edge.
\end{itemize}

Based on this assumption, the adjacency matrix can be calculated using equation (\ref{eq:adjmatrix}).
\begin{equation}
    \label{eq:adjmatrix}
    A^l=\|V^l\|_F\cdot\|V^l\|_F^T
\end{equation}
where $\|\cdot\|_F$ represents the Frobenius norm, which is used to ensure numerical stability while introducing self-loops. $\cdot$ represents matrix multiplication, $^T$ represents matrix transpose.

\subsubsection{GCN fusion algorithm}
After obtaining the adjacency matrix, GCN needs to normalize the adjacency matrix using a degree matrix, which can be obtained by the formula (\ref{eq:degreematrix}).
\begin{equation}
    \label{eq:degreematrix}
    D^l=\begin{bmatrix}
        \sum_{i}|A^l_{i,1}| & 0 & \ldots & 0\\
        0 & \sum_{i}|A^l_{i,2}| & \ldots & 0\\
        \vdots & \vdots & \ddots & \vdots\\
        0 & 0 & 0 & \sum_{i}|A^l_{i,L/16}|
    \end{bmatrix}^{-\frac{1}{2}}
\end{equation}
where $|\cdot|$ represents an absolute value operation. 
The degree matrix $D^l\in\mathbb{R}^{\frac{L}{16}\times\frac{L}{16}}$ is a diagonal matrix.

After obtaining the degree matrix, the adjacency matrix can be normalized using equation (\ref{eq:normadj}).
\begin{equation}
    \label{eq:normadj}
    \tilde{A}^l=D^l  A^l (D^l)^T
\end{equation}

Next, we use a two-layer GCN network to fuse node features, as shown in equation (\ref{eq:gcnfusion}).
\begin{equation}
    \label{eq:gcnfusion}
    \tilde{V}^l=\tilde{A}^l\sigma(\tilde{A}^lV^lW^l_1)W^l_2
\end{equation}
where $W_1^l$ and $W_2^l$ represent the weights of the network, respectively.
$\sigma(\cdot)$ denotes a nonlinear activation function (GeLU).

\subsubsection{GAT fusion algorithm}
GAT needs to calculate the attention coefficient of different graph nodes based on neighboring nodes, as shown in the formula (\ref{eq:gatscore}). 
\begin{equation}
    \label{eq:gatscore}
    e_{ij}=a([V_i^lW_3\|V^l_jW_3]), j\in\mathcal{N}_i
\end{equation}
where $W\in\mathbb{R}^{C\times 2C}$ is a linear projection for feature augmentation, $[\cdot\|\cdot]$ denotes a feature concatenating operation. $a(\cdot):\mathbb{R}^{4C}\rightarrow\mathbb{R}$ is a score function which is a single-layer FNN.

Then the attention score can be calculated by formula (\ref{eq:gatattn}).
\begin{equation}
    \label{eq:gatattn}
    \alpha_{ij}=\frac{\exp(\text{LeakyReLU}(e_{ij}))}{\sum_{k\in\mathcal{N}_i}\exp(\text{LeakyReLU}(e_{ik}))}
\end{equation}

The output feature can be aggregated by using the formula (\ref{eq:gatagg}).
\begin{equation}
    \label{eq:gatagg} \tilde{V}^l_{i}(H)=\| _{h=1}^H\sigma\left(\sum_{j\in\mathcal{N}_i} \alpha_{ij}^h V^l_jW_3^h\right)
\end{equation}
where $\|_{h=1}^H$ denotes the multi-head attention, $H$ denotes the number of head. $\sigma(\cdot)$ is an Exponential Linear Unit (ELU) activation function. 

GNN can fully learn structured features that match the user's intent to build robust click control.
In the experimental section, we will compare the performance of GCN vs. GAT.

\subsection{Dual cross-attention for improving segmentation control robustness}
The structured interaction features obtained using GNN can be used to guide the ViT features, making them better meet the user's interaction intentions.

To fully utilize user clicks, we use the formula (\ref{eq:cat4query}) to concatenate the positive and negative interaction features as the final guiding feature.
\begin{equation}
    \label{eq:cat4query}
    x_q = [\tilde{\mathcal{V}}^0+\gamma^0, \tilde{\mathcal{V}}^1+\gamma^1]
\end{equation}
where $x_q\in\mathbb{R}^{B\times\frac{L}{8}\times C}$ denotes the final guiding feature. $\gamma^l\in\mathbb{R}^{1\times 1\times C}$ represents the token used to distinguish between positive and negative clicks.

We use dual cross attention to ensure a better fusion of interaction features and ViT features, as shown in the formula (\ref{eq:twowaycross}).
\begin{equation}
    \label{eq:twowaycross}
    \begin{split}
        \tilde{x}_q &= x_{q} + \text{cross-attn}(x_{q}, x_{f})\\
        \tilde{x}_{f} &= x_{f} + \text{cross-attn}(x_{f}, \tilde{x}_q)\\
        \hat{x}_{f} &= \tilde{x}_{f} + \text{FNN}(\tilde{x}_{f})
    \end{split}
\end{equation}
where $\text{cross-attn}(\cdot,\cdot)$ represents cross attention operation, $\text{FNN}(\cdot)$ represents FNN. $\hat{x}_{f}$ will be used for target object segmentation.

\section{Experiment}

In this chapter, we detail the basic settings, training, and interactive validation strategy of the proposed algorithm. It's compared with existing SOTA algorithms in accuracy and speed, and its effectiveness is further confirmed through ablation studies.
The proposed algorithm is referred to as \textbf{SCC} in subsequent sections.

\subsection{Experimental Configuration}

\textbf{Model selection.}
We selected three representative algorithms to validate the effectiveness of \textbf{SCC} as a universal click control model.
These algorithms include Vision Transformer-based \cite{dosovitskiy2020image} SimpleClick \cite{liu2022simpleclick}, the Segformer-based \cite{xie2021segformer} FocalClick \cite{chen2022focalclick}, and recent popular Segment Anything (SAM) \cite{kirillov2023segment,sam_hq} algorithms.

\textbf{Training settings.}

The training data size is $448\times 448$, iterative learning \cite{9897365} and click sample strageties are adopted for click simulation, maximum clicks is  24 with 0.8 decay probability.
AdamW optimizer is used, with $\beta_1=0.9, \beta_2=0.999$. Each epoch comprises 30,000 samples, totaling 230 epochs. The initial learning rate is $5\times 10^{-6}$, reducing by 10x at epochs 50 and 70. Training on six Nvidia RTX 3090 GPUs.

\textbf{Evaluation strategy.}
The widely used evaluation techniques \cite{chen2021conditional, 9156403, 9897365} usually sample each point from the largest error locations in previous predictions, aiming for an Intersection Over Union (IOU) close to the target.

However, we found in practice that users often have difficulty identifying the maximum error point, and the locations they click on often exhibit significant randomness. Therefore, this strategy fails to effectively reflect the performance of the algorithm in practical applications.

Therefore, we propose a new click generation strategy, which randomly samples click positions from the set of top-$k$ locations, thereby better reflecting the performance of the algorithm in practical applications. In this article, we set $k=50$ to better simulate real user interaction

For fairness comparison, both strategies will be employed.

The evaluation concludes upon reaching the desired IOU or maximum click count. 
The standard Number of Clicks (NoC) metric and the Number of Failures (NoF) metric are adopted \cite{chen2022focalclick}.

\subsection{Comparison with State-of-the-Art}
This section compares the proposed algorithm with current SOTA in accuracy and complexity, further analyzing \textbf{SCC}'s effectiveness.

\noindent\textbf{1) Overall performance on mainstream benchmarks. }

All algorithms are trained on the COCO+LVIS dataset. 
We first compared the overall performance using the "largest error" point sampling strategy, and the results on key benchmarks are shown in Table \ref{tab:sotacompar}.

\begin{table*}[!t]
\centering
\caption{Evaluation results on GrabCut, Berkeley, SBD, DAVIS, COCO Mval, and Pascal VOC. 'NoC 85/90' denotes the average Number of Clicks required the get IoU of 85/90\%. All methods are trained on COCO\cite{lin2014microsoft} and LVIS\cite{gupta2019lvis} datasets. \textbf{Bold font}: best performance}
\label{tab:sotacompar}
\scalebox{0.9}{
\begin{tabular}{ccccccccccc}
\hline
 &  & GrabCut \cite{rother2004grabcut} & Berkeley \cite{mcguinness2010comparative} & \multicolumn{2}{c}{SBD \cite{MCGUINNESS2010434}} & DAVIS \cite{perazzi2016benchmark} & \multicolumn{2}{c}{COCO MVal \cite{10.1007/978-3-319-10602-1_48}} & \multicolumn{2}{c}{PascalVOC \cite{everingham2009pascal}} \\
\multirow{-2}{*}{Arch} & \multirow{-2}{*}{Method} & NoC 90 & NoC 90 & NoC 85 & NoC 90 & NoC 90 & NoC 85 & NoC 90 & NoC 85 & NoC 90 \\ \hline
 & f-BRS-B-hrnet32\cite{9156403} & 1.69 & 2.44 & 4.37 & 7.26 & 6.50 & - & - & - & - \\
 & RITM-hrnet18s\cite{9897365} & 1.68 & 2.60 & 4.25 & 6.84 & 5.98 & - & 3.58 & 2.57 & - \\
 & RITM-hrnet32\cite{9897365} & 1.56 & 2.10 & 3.59 & 5.71 & 5.34 & 2.18 & 3.03 & 2.21 & 2.59 \\
\multirow{-4}{*}{CNN} & EdgeFlow-hrnet18\cite{Hao_2021_ICCV} & 1.72 & 2.40 & - & - & 5.77 & - & - & - & - \\ \hline
 & FocalClick-segformer-B3-S2\cite{chen2022focalclick} & 1.68 & 1.71 & 3.73 & 5.92 & 5.59 & 2.45 & 3.33 & 2.53 & 2.97 \\ \cline{2-11} 
 & \cellcolor[HTML]{EFEFEF}FocalClick-segformer-B3-S2+GCN & \cellcolor[HTML]{EFEFEF} 1.52 & \cellcolor[HTML]{EFEFEF} 1.61 & \cellcolor[HTML]{EFEFEF} 3.78 & \cellcolor[HTML]{EFEFEF} 5.96 & \cellcolor[HTML]{EFEFEF} 4.41 & \cellcolor[HTML]{EFEFEF} 2.54 & \cellcolor[HTML]{EFEFEF} 3.49 & \cellcolor[HTML]{EFEFEF} 2.56 & \cellcolor[HTML]{EFEFEF} 3.02 \\
 & \cellcolor[HTML]{EFEFEF}FocalClick-segformer-B3-S2+GAT & \cellcolor[HTML]{EFEFEF} 1.52 & \cellcolor[HTML]{EFEFEF} 1.48 & \cellcolor[HTML]{EFEFEF} 3.66 & \cellcolor[HTML]{EFEFEF} 5.91 & \cellcolor[HTML]{EFEFEF} 3.77 & \cellcolor[HTML]{EFEFEF} 2.34 & \cellcolor[HTML]{EFEFEF} 3.16 & \cellcolor[HTML]{EFEFEF} 2.42 & \cellcolor[HTML]{EFEFEF} 2.87\\\cline{2-11}
 & HQ-SAM-ViT-H\cite{sam_hq} & 1.84 & 2.00 & 6.23 & 9.66 & 5.58 & 3.81 & 5.94 & 2.50 & 2.93 \\
& HQ-SAM-ViT-B\cite{sam_hq} & 2.28 & 4.26 & 11.11 & 13.57 & 8.19 & 6.41 & 9.54 & 4.86 & 6.27  \\\cline{2-11}
  & \cellcolor[HTML]{EFEFEF}HQ-SAM-ViT-B+GCN & \cellcolor[HTML]{EFEFEF} 2.34 & \cellcolor[HTML]{EFEFEF} 3.74 & \cellcolor[HTML]{EFEFEF} 9.46 & \cellcolor[HTML]{EFEFEF} 12.99 & \cellcolor[HTML]{EFEFEF} 7.38  & \cellcolor[HTML]{EFEFEF} 5.87 & \cellcolor[HTML]{EFEFEF} 8.84 & \cellcolor[HTML]{EFEFEF} 5.80 & \cellcolor[HTML]{EFEFEF} 8.21 \\
  Transformer & \cellcolor[HTML]{EFEFEF}HQ-SAM-ViT-B+GAT & \cellcolor[HTML]{EFEFEF} 2.02 & \cellcolor[HTML]{EFEFEF} 3.87 & \cellcolor[HTML]{EFEFEF} 9.11 & \cellcolor[HTML]{EFEFEF} 12.76 & \cellcolor[HTML]{EFEFEF} 7.16 & \cellcolor[HTML]{EFEFEF} 5.54 & \cellcolor[HTML]{EFEFEF} 8.33 & \cellcolor[HTML]{EFEFEF} 4.50 & \cellcolor[HTML]{EFEFEF} 5.58 \\ \cline{2-11} 
 & SimpleClick-ViT-B\cite{liu2022simpleclick} & 1.48 & 1.97 & 3.43 & 5.62 & 5.06 & 2.18 & 2.92 & 2.06 & 2.38 \\
 & SimpleClick-ViT-L\cite{liu2022simpleclick} & 1.46 & 2.33 & 2.69 & 4.46 & 5.39 & - & - & 1.95 & 2.30 \\ \cline{2-11} 
 & \cellcolor[HTML]{EFEFEF}SimpleClick-ViT-B+GCN & \cellcolor[HTML]{EFEFEF} 1.34 & \cellcolor[HTML]{EFEFEF} 1.60 & \cellcolor[HTML]{EFEFEF} 3.11 & \cellcolor[HTML]{EFEFEF} 5.00 & \cellcolor[HTML]{EFEFEF} 4.58 & \cellcolor[HTML]{EFEFEF} 2.17 & \cellcolor[HTML]{EFEFEF} 2.95 & \cellcolor[HTML]{EFEFEF} 1.71 & \cellcolor[HTML]{EFEFEF} 1.94 \\
 & \cellcolor[HTML]{EFEFEF}SimpleClick-ViT-B+GAT & \cellcolor[HTML]{EFEFEF} 1.34 & \cellcolor[HTML]{EFEFEF} 1.41 & \cellcolor[HTML]{EFEFEF} 2.98& \cellcolor[HTML]{EFEFEF} 4.88 & \cellcolor[HTML]{EFEFEF} 4.39 & \cellcolor[HTML]{EFEFEF} 2.21 & \cellcolor[HTML]{EFEFEF} 3.04 & \cellcolor[HTML]{EFEFEF} 1.69 & \cellcolor[HTML]{EFEFEF} 1.92 \\
 & \cellcolor[HTML]{EFEFEF}SimpleClick-ViT-L+GCN & \cellcolor[HTML]{EFEFEF} 1.32 & \cellcolor[HTML]{EFEFEF} 1.34 & \cellcolor[HTML]{EFEFEF} 2.60 & \cellcolor[HTML]{EFEFEF} 4.28 & \cellcolor[HTML]{EFEFEF} 4.03 & \cellcolor[HTML]{EFEFEF}2.01 & \cellcolor[HTML]{EFEFEF} 2.77 & \cellcolor[HTML]{EFEFEF} 1.58 & \cellcolor[HTML]{EFEFEF} 1.76\\
& \cellcolor[HTML]{EFEFEF}SimpleClick-ViT-L+GAT & \cellcolor[HTML]{EFEFEF} 1.34 & \cellcolor[HTML]{EFEFEF} 1.43 & \cellcolor[HTML]{EFEFEF} 2.52 & \cellcolor[HTML]{EFEFEF} 4.21 & \cellcolor[HTML]{EFEFEF} 3.95 & \cellcolor[HTML]{EFEFEF} 1.99 & \cellcolor[HTML]{EFEFEF} 2.73 & \cellcolor[HTML]{EFEFEF} 1.56 & \cellcolor[HTML]{EFEFEF} 1.73 \\ \hline
\end{tabular}
}
\end{table*}
As shown in Table \ref{tab:sotacompar}, the proposed \textbf{SCC}-based algorithms can significantly improve performance in different benchmarks.
Especially the GAT-based algorithm demonstrate better performence than GCN.

It can be seen that both GCN and GAT in \textbf{SCC} show significant improvements in different algorithms and evaluation metrics, which validates the effectiveness of \textbf{SCC} in enhancing algorithm performance.


To evaluate effectiveness of \textbf{SCC} in real applications, we also compared the overall performance using the "top-k error" point sampling strategy. 
Due to layout and space limitations, we only adopted this strategy in algorithm Simplick and evaluated on SBD and DAVIS.
In addition, due to the randomness in the selection of top-k clicks, we use the average of ten evaluations as the final result, as shown in Table \ref{tab:topksota}.

\begin{table}[!htb]
\centering
\caption{In actual use, the effectiveness of SCC performance is demonstrated. The number in parentheses represents the total number of samples in the benchmark }
\label{tab:topksota}
\scalebox{0.85}{
\begin{tabular}{cccccc}
\hline
\multirow{2}{*}{Method} & \multicolumn{2}{c}{DAVIS} & \multicolumn{3}{c}{SBD} \\
 & NoC 90 & NoF 90 & NoC 85 & NoC 90 & NoF 90 \\ \hline
SimpleClick-ViT-B & 5.10 & 50/(345) & 3.81 & 6.29 & 1056/(6671) \\ \cline{1-6}
\cellcolor[HTML]{EFEFEF} +GAT & \cellcolor[HTML]{EFEFEF} 4.67 &  \cellcolor[HTML]{EFEFEF} 48/(345)& \cellcolor[HTML]{EFEFEF} 3.54 & \cellcolor[HTML]{EFEFEF} 5.71 & \cellcolor[HTML]{EFEFEF} 918/(6671)\\ \hline
HQ-SAM-ViT-B & 8.19 &  117/(345) & 10.14 & 13.57 &  4123/(6671) \\\cline{1-6}
\cellcolor[HTML]{EFEFEF} +GAT & \cellcolor[HTML]{EFEFEF} 7.16  & \cellcolor[HTML]{EFEFEF} 91/(345) & \cellcolor[HTML]{EFEFEF} 9.11 & \cellcolor[HTML]{EFEFEF} 12.76 & \cellcolor[HTML]{EFEFEF} 3789/(6671) \\ \hline
FocalClick-B3-S2 & 4.88 &  26/(345) & 4.0 & 6.31 &  1024/(6671) \\\cline{1-6}
\cellcolor[HTML]{EFEFEF} +GAT & \cellcolor[HTML]{EFEFEF} 3.77  & \cellcolor[HTML]{EFEFEF} 31/(345) & \cellcolor[HTML]{EFEFEF} 3.66 & \cellcolor[HTML]{EFEFEF} 5.91 & \cellcolor[HTML]{EFEFEF} 870/(6671) \\ \hline
\end{tabular}
}
\end{table}

To simplify the experiments, we only tested \textbf{GAT} in real-world scenarios. It can be observed that in simulated scenarios reflecting real user usage, the proposed algorithm still effectively improves the performance of the baseline algorithm in various aspects.

Furthermore, the experimental results in Table \ref{tab:topksota} do not show significant performance differences compared to those in Table \ref{tab:sotacompar} on small-scale datasets. However, on large-scale datasets like SBD, conventional evaluation methods may not adequately reflect the algorithm's actual performance. From this perspective, the performance comparison experiments under real-world scenarios proposed in this paper can provide more robust performance benchmarks for individuals using interactive segmentation algorithms in practice.

\textbf{2) Complexity Analysis}

We evaluate the complexity of algorithms by comparing their parameters count, floating point operations per second (FLOPs), and inference speed. The experimental results are listed in Table \ref{tab:speed_tab}.

\begin{table}[!h]
\centering
\caption{The analysis of the computational complexity of SCC, including parameter count (Params) and floating point operations (FLOPs) }
\label{tab:speed_tab}
\scalebox{0.95}{
\begin{tabular}{cccc}
\hline
Model Type & SCC & Params (MB) & FLOPs (G) \\ \hline
\multirow{2}{*}{HQ-SAM \& SimpleClick} & +GCN & +10.564 & +13 \\
 & +GAT & +8.275 & +10.333 \\ \hline
\multirow{2}{*}{FocalClick} & +GCN & +6.955 & +0.539 \\
 & +GAT & +3.867 & +0.511 \\ \hline
\end{tabular}
}
\end{table}

The experimental results indicate that GCN requires slightly more parameters compared to GAT, while the parameter count of both models does not exhibit a significant increase compared to the parameter count of the original algorithm.

For instance, the parameter count of FocalClick-B3-S2 is 45.66MB, where the parameter count of GAT in four stages only accounts for 8\% of it. For HQ-SAM-ViT-B, with a parameter count of 637.23MB, the parameter count of GAT in two stages only constitutes 1.3\% of it. Similarly, for SimpleClick-ViT-B, with a parameter count of 84.49MB, the parameter count of GAT in two stages only occupies 9.8\% of it.

In summary, the proposed algorithm significantly improves algorithm performance without a noticeable increase in algorithm complexity.

\subsection{Ablation Study}
The ablation study results are shown in Table {\ref{tab:sotacompar}}. 
Which can be seen that the proposed \textbf{SCC} can significintly improve the performance in reducing the click times.

The performance of GCN fails to improve click performance on certain datasets, mostly concentrated in SBD, PascalVOC, and COCO MVal datasets. This indicates that GCN models based on fixed adjacency matrices may not effectively model user interactions in large-scale complex scenarios. In contrast, the proposed GAT model consistently improves algorithm performance across almost all performance metrics. This validates that the mechanism based on local node attention may be more suitable for modeling user clicks in complex scenarios.

\subsection{Performance for Mask Correction}

The mask correction task involves adjusting given masks with typical IOU values ranging from 0.7 to 0.85. 
Since SAM does not directly utilize the mask from the previous frame, we have not evaluated its performance on mask correction tasks.
To simplify the experimental complexity, we directly utilized \textbf{GAT} for comparison.
The performance of the proposed algorithm in improving baseline on this task (DAVIS-585 \cite{chen2022focalclick}) is shown in Table \ref{tab:maskcorrection}.

\begin{table}[!h]
\setlength{\tabcolsep}{3pt}
\centering
\caption{Quantitative results on DAVIS-585 benchmark. The metrics ‘NoC’ and ‘NoF’ mean the average Number of Clicks required and the Number of Failure examples for the target IOU.}
\label{tab:maskcorrection}
\scalebox{0.76}{
\begin{tabular}{cccc|ccc}
\hline
 & \multicolumn{3}{c|}{DAVIS585-SP} & \multicolumn{3}{c}{DAVIS585-ZERO} \\ 
 & NoC85 & NoC90 & NoF85 & NoC85 & NoC90 & NoF85 \\ \hline
RITM-hrnet18s\cite{9897365} & 3.71 & 5.96 & 49 & 5.34 & 7.57 & 52 \\
RITM-hrnet32\cite{9897365} & 3.68 & 5.57 & 46 & 4.74 & 6.74 & 45 \\
SimpleClick-ViT-B\cite{liu2022simpleclick} & 2.24 & 3.10 & 25 & 4.06 & 5.83 & 42 \\\hline
\cellcolor[HTML]{EFEFEF} +GAT & \cellcolor[HTML]{EFEFEF} 2.11$_{\textcolor{red}{\uparrow{5.3\%}}}$ & \cellcolor[HTML]{EFEFEF} 2.83$_{\textcolor{red}{\uparrow{8.7\%}}}$ & \cellcolor[HTML]{EFEFEF} 23 &  \cellcolor[HTML]{EFEFEF} 3.78$_{\textcolor{red}{\uparrow{6.9\%}}}$ &\cellcolor[HTML]{EFEFEF} 5.33$_{\textcolor{red}{\uparrow{8.6\%}}}$ &  \cellcolor[HTML]{EFEFEF} 42\\ \hline
SimpleClick-ViT-L\cite{liu2022simpleclick} & 1.81 & 2.57 & 25 & \textbf{3.39} & \textbf{4.88} & 36 \\
FocalClick-hrnet32-S2\cite{chen2022focalclick} & 2.32 & 3.09 & 28 & 4.77 & 6.84 & 48 \\
FocalClick-B3-S2\cite{chen2022focalclick} & 2.00 & 2.76 & 22 & 4.06 & 5.89 & 43 \\ \hline
\cellcolor[HTML]{EFEFEF} +GAT & \cellcolor[HTML]{EFEFEF} 2.00 & \cellcolor[HTML]{EFEFEF} 2.53$_{\textcolor{red}{\uparrow{8.3\%}}}$ & \cellcolor[HTML]{EFEFEF} 20 &  \cellcolor[HTML]{EFEFEF} 3.77$_{\textcolor{red}{\uparrow{7.1\%}}}$ &\cellcolor[HTML]{EFEFEF} 5.42$_{\textcolor{red}{\uparrow{7.9\%}}}$ &  \cellcolor[HTML]{EFEFEF} 42\\ \hline
\end{tabular}
}
\end{table}

The experimental results indicate that the proposed algorithm achieved an average improvement of 8\% in the performance metric of NoC 90 for the mask correction task.
Simultaneously, the proposed algorithm also exhibited a significant improvement over the baseline algorithm in the NoF metric.
These results validate the effectiveness of the proposed algorithm in mask correction.

\begin{figure*}[!htp]
\centering
\includegraphics[scale=0.37]{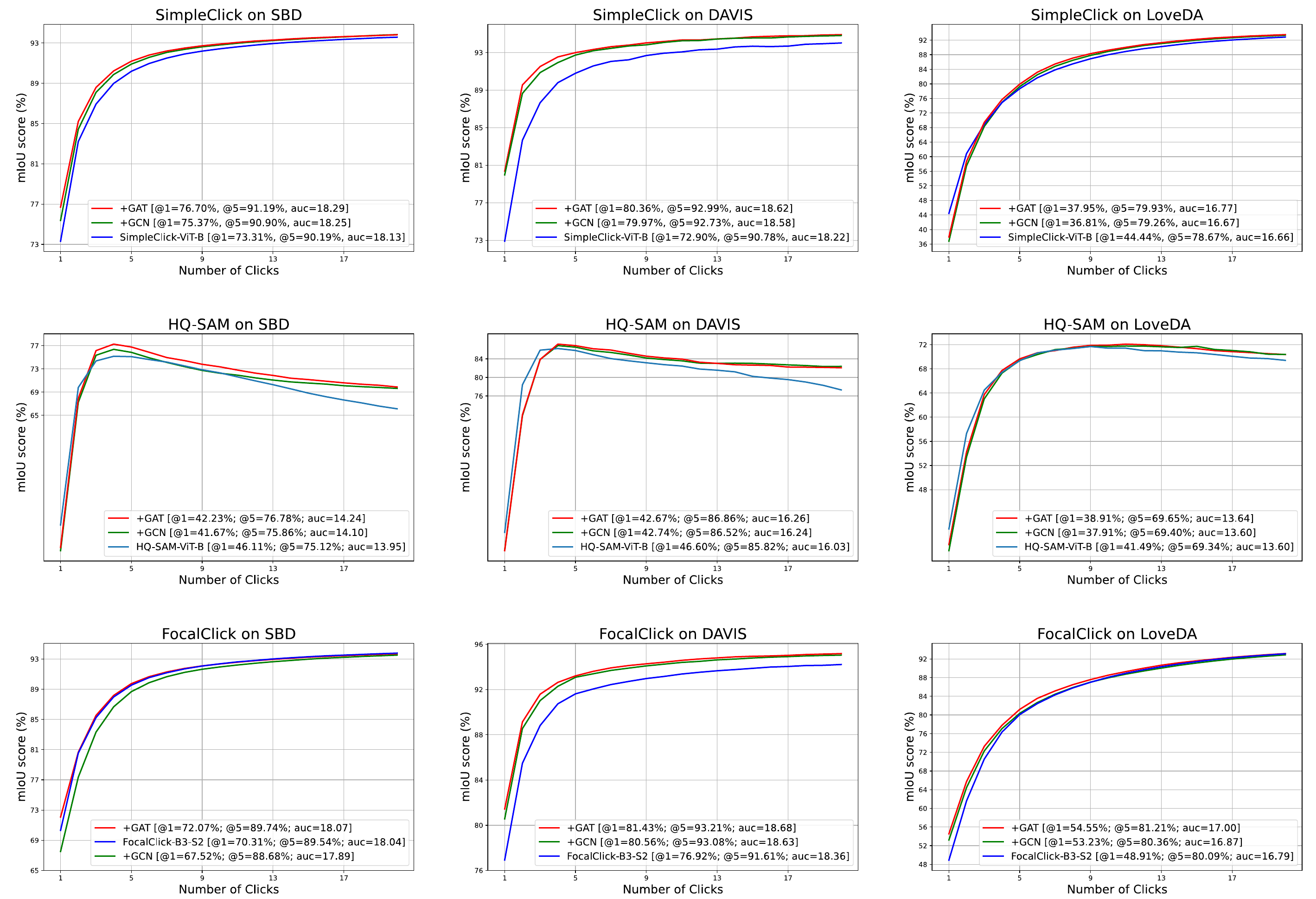}
\caption{Trade off between mIoU and number of clicks}
\label{fig:controlauc}
\end{figure*}
\subsection{Click Control Evaluation}

To evaluate the effectiveness of the proposed algorithm in enhancing user clicks for adjusting model outputs, we assessed the relationship between the number of clicks and mIoU (mean Intersection over Union), and plotted the curve illustrating their relationship.

In our experiment, we utilized 20 clicks and employed two large-scale datasets, SBD (containing 6671 instances) and LoveDA (containing 1666 instances), as well as a finely annotated dataset, DAVIS (containing 345 instances).
The experimental results are depicted in Figure \ref{fig:controlauc}.

Analyzing the experimental results, we observe that the proposed algorithm effectively enhances the accuracy of the baseline model after a single click across all three datasets, particularly demonstrating significant improvement on finely annotated datasets like DAVIS.

However, in the initial clicks, the improvement magnitude of the proposed algorithm on HQ-SAM-ViT-B is significantly lower compared to FocalClick and SimpleClick. This is because we conducted full-parameter training for FocalClick and SimpleClick, while only performing partial parameter fine-tuning on HQ-SAM-ViT-B.

It is worth noting that although the overall accuracy of HQ-SAM-ViT-B decreases with the user's clicks increasing, the proposed algorithm exhibits a clear trend of suppressing performance degradation.
This result effectively demonstrates the effectiveness of the proposed algorithm in enhancing user click control.

\begin{figure*}[!t]
\centering
\includegraphics[scale=0.53]{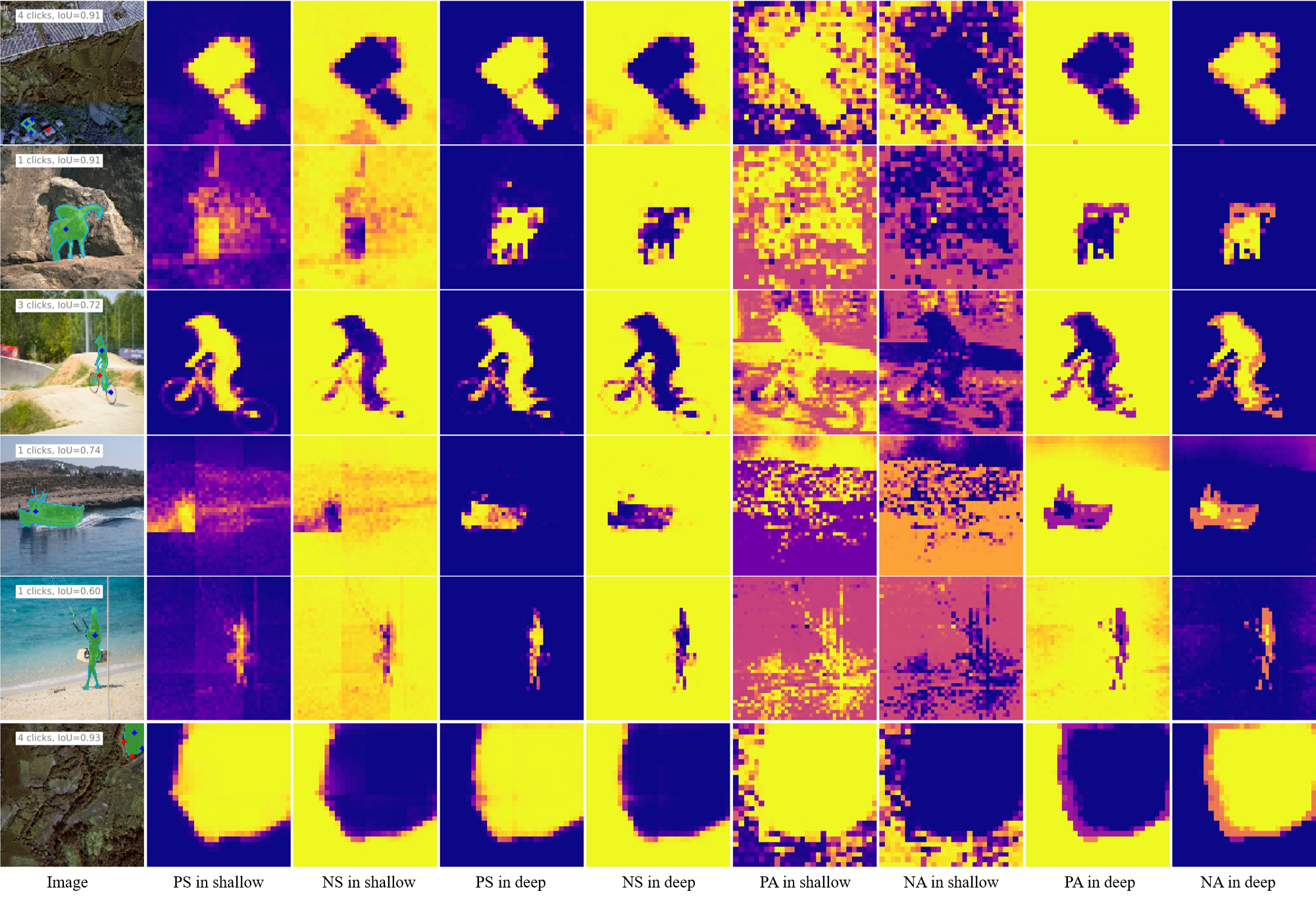}
\caption{Analysis of intermediate variables in the GAT model. \textbf{PS} represents the scores of tokens similar to positive clicks, \textbf{NS} represents the scores of tokens similar to negative clicks, \textbf{PA} represents attention calculated from tokens filtered by \textbf{PS}, \textbf{NA} represents attention calculated from tokens filtered by \textbf{NS}. \textbf{shallow} represents shallow features, while deep represents \textbf{deep} features}
\label{fig:quali_analysis}
\end{figure*}

To quantitatively analyze the effectiveness of the proposed algorithm in enhancing control performance, we have listed the corresponding AUC performance improvement metrics on three datasets, as shown in Table \ref{tab:control_tab}.

\begin{table}[!htb]
\centering
\caption{Performance evaluation of click control based on area under the curve}
\label{tab:control_tab}
\scalebox{1.0}{
\begin{tabular}{cccc}
\hline
Method & LoveDA & DAVIS  & SBD \\ \hline
SimpleClick-ViT-B & 16.66 & 18.22 & 18.13 \\ \cline{1-4}
\cellcolor[HTML]{EFEFEF} SimpleClick+GCN & \cellcolor[HTML]{EFEFEF} 16.67$_{\textcolor{red}{\uparrow{0.06\%}}}$& \cellcolor[HTML]{EFEFEF} 18.58$_{\textcolor{red}{\uparrow{2\%}}}$ & \cellcolor[HTML]{EFEFEF} 18.25$_{\textcolor{red}{\uparrow{0.6\%}}}$ \\
\cellcolor[HTML]{EFEFEF} SimpleClick+GAT & \cellcolor[HTML]{EFEFEF} 16.77$_{\textcolor{red}{\uparrow{0.6\%}}}$ & \cellcolor[HTML]{EFEFEF} 18.62$_{\textcolor{red}{\uparrow{2.2\%}}}$ & \cellcolor[HTML]{EFEFEF} 18.29$_{\textcolor{red}{\uparrow{0.8\%}}}$ \\ \hline
FocalClick & 16.79 & 18.36 & 18.04 \\ \cline{1-4}
\cellcolor[HTML]{EFEFEF} FocalClick+GCN & \cellcolor[HTML]{EFEFEF} 16.87$_{\textcolor{red}{\uparrow{0.5\%}}}$ & \cellcolor[HTML]{EFEFEF} 18.63$_{\textcolor{red}{\uparrow{1.4\%}}}$ & \cellcolor[HTML]{EFEFEF} 17.89 \\
\cellcolor[HTML]{EFEFEF} FocalClick+GAT & \cellcolor[HTML]{EFEFEF} 17.00$_{\textcolor{red}{\uparrow{1.3\%}}}$ & \cellcolor[HTML]{EFEFEF} 18.68$_{\textcolor{red}{\uparrow{1.7\%}}}$ & \cellcolor[HTML]{EFEFEF} 18.07$_{\textcolor{red}{\uparrow{0.16\%}}}$ \\ \hline
HQ-SAM-ViT-B & 13.60 & 16.03 & 13.95 \\\cline{1-4}
\cellcolor[HTML]{EFEFEF} HQ-SAM+GCN & \cellcolor[HTML]{EFEFEF} 13.60 &  \cellcolor[HTML]{EFEFEF} 16.24$_{\textcolor{red}{\uparrow{1.3\%}}}$ & \cellcolor[HTML]{EFEFEF} 14.10$_{\textcolor{red}{\uparrow{1.0\%}}}$\\
\cellcolor[HTML]{EFEFEF} HQ-SAM+GAT & \cellcolor[HTML]{EFEFEF} 13.64$_{\textcolor{red}{\uparrow{0.3\%}}}$&  \cellcolor[HTML]{EFEFEF} 16.26$_{\textcolor{red}{\uparrow{1.4\%}}}$ & \cellcolor[HTML]{EFEFEF} 14.24$_{\textcolor{red}{\uparrow{2\%}}}$\\ \hline
\end{tabular}
}
\end{table}

The experimental results indicate that the improvement of the proposed algorithm on large-scale datasets like LoveDA and SBD generally ranges from 0.06\% to 1.3\%, while on DAVIS, the improvement is more significant, ranging from 1.3\% to 2.2\%.

Although on average, the improvement of the proposed algorithm on large-scale datasets is relatively small, as depicted in Figure \ref{fig:controlauc}, we observe a significant performance improvement on more practically meaningful metrics such as performance after one click and five clicks. 
These experimental results demonstrate that the proposed algorithm can effectively enhance the efficiency of user control over network outputs while also improving model output accuracy.

\subsection{Quantitative analysis}
In this section, we first analyze the effectiveness of the proposed algorithm itself, namely the degree of match between the results output by different components in practical computations and the expected outcomes. Then, we present a representative scenario of click failure faced by the baseline algorithm, and analyze the effectiveness of the proposed algorithm in addressing this scenario.

To evaluate the effects of the GNN module in affecting network decision, we analysised the similarity and attention calculation during click control. Some results are visualized in Figure \ref{fig:quali_analysis}.

It can be observed that GAT effectively computes the similarity between user-clicked tokens and the target, which is crucial for subsequent computations. Furthermore, it can be noted that the similarity computed from deep features exhibits higher quality, whereas the similarity computed from shallow features tends to be relatively lower. This trend is further reflected in the attention calculation of deep and shallow features. The attention obtained from shallow-scored structured tokens is coarser compared to that obtained from deep-scored structured tokens.

The visualization results of these intermediate variables indicate that the proposed algorithm effectively realizes the expected assumptions, thereby guiding the model's attention layers to focus more on the target regions that align with user intent. Additionally, the interaction between shallow and deep features brings about significant feature diversity, enabling the algorithm to better capture precise user intent from multiple perspectives.

\begin{figure*}[!t]
\centering
\includegraphics[scale=0.51]{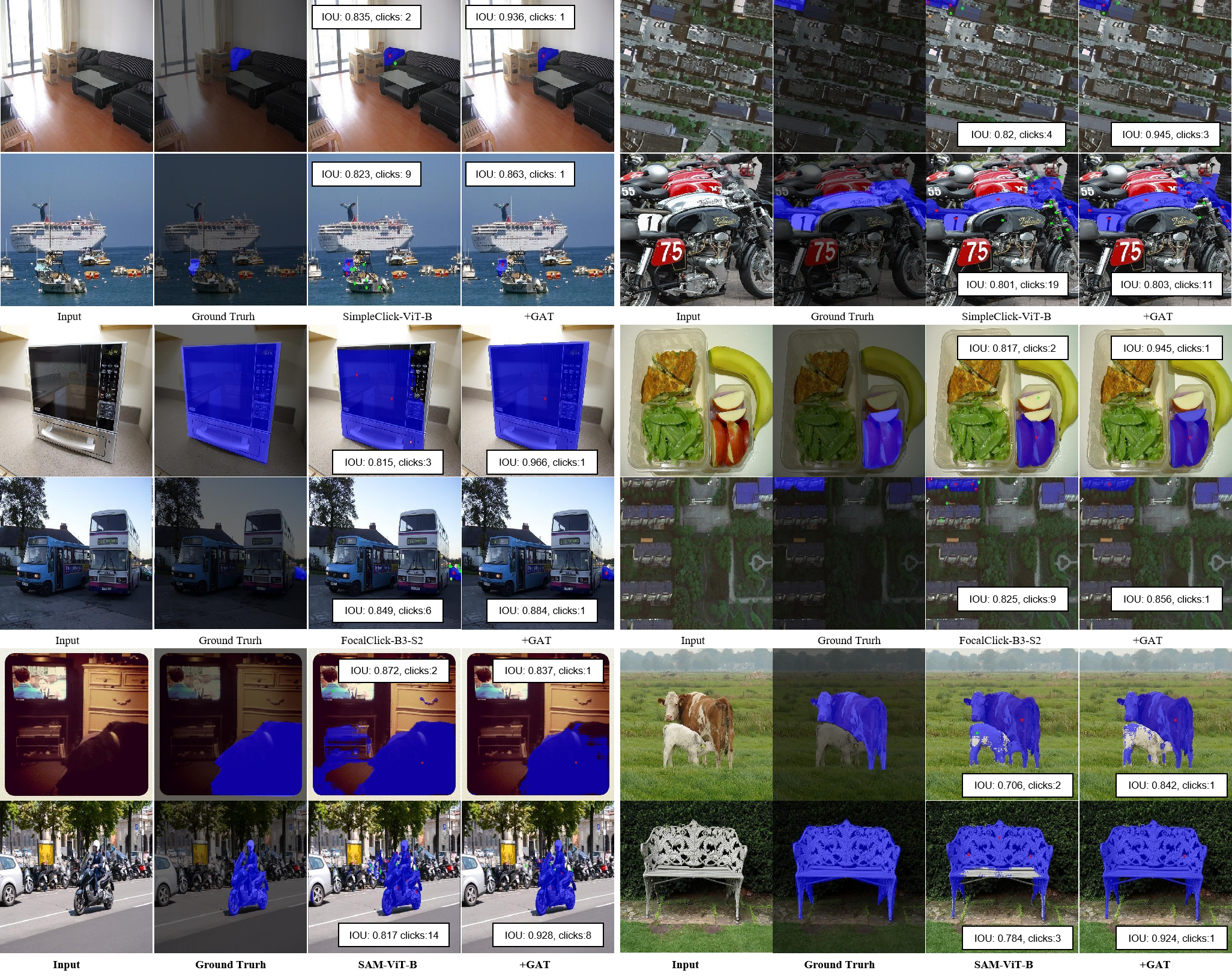}
\caption{The quantitative analysis results of the proposed algorithm and the baseline algorithm in certain scenarios.  The red dots represent positive clicks, while the green dots represent negative clicks. We compared the performance before and after adding GAT for FocalClick, SimpleClick, and SAM respectively. Each algorithm corresponds to four samples. The figure illustrates the IOU at different numbers of clicks }
\label{fig:quali}
\end{figure*}

Finally, to intuitively evaluate the effectiveness of the proposed algorithm in addressing the issue of click loss, we further quantitatively analyzed the performance comparison between the proposed algorithm and the baseline algorithm in some typical scenarios. The experimental results are shown in Figure \ref{fig:quali}.

It can be observed that GAT demonstrates a significant performance improvement compared to the baseline algorithm in the "one-click" scenario for all three algorithms.
This is because GAT filters out a larger range of user intent through similarity selection, ensuring that a single click can have a similar impact to multiple clicks.

Furthermore, compared to the baseline algorithm, GAT also exhibits better click control ability in scenarios involving multiple clicks.
In some cases, eliminating erroneous regions in the output requires only a few clicks. In contrast, the baseline algorithm requires more negative clicks to eliminate undesired outputs.
This trend is particularly pronounced in the SAM model.

Overall, through the analysis of GNN intermediate variables and the effectiveness of the proposed algorithm in improving baseline performance, we found a high degree of consistency between the GNN intermediate variables and our expectations within our framework. Additionally, the performance of the proposed algorithm also demonstrated better performance than the baseline algorithm in both single-click and multiple-click scenarios. These experimental results validate the effectiveness of the proposed algorithm.
\subsection{Generalization Evaluation on Remote Sensing Images}
In our study, all compared algorithms are trained on COCO-LVIS. However, COCO-LVIS' focus on common image types and scenes means it lacks generalization in some special visual data like remote sensing images.

In this section, we test all methods on the remote sensing dataset LoveDA \cite{loveda} and Rice \cite{rice}, as shown in Table \ref{tab:rs_compar}.

\begin{table}[!h]
\caption{Comparison of algorithm generalization performance on LoveDA \cite{loveda} and Rice \cite{rice}. None of the algorithms were trained on remote sensing images}
\label{tab:rs_compar}
\centering
\scalebox{1.0}{
\begin{tabular}{m{2.2cm} ccccc}
\hline
\multicolumn{5}{c}{LoveDA \cite{loveda}} \\
 & NoC 80 & NoC 85 & NoC 90 & NoF 90 \\ \hline
HQ-SAM-ViT-B & 11.17 & 14.70 & 17.73  & 1409/(1666)\\\cline{1-5}
+GAT\cellcolor[HTML]{EFEFEF} & \cellcolor[HTML]{EFEFEF}11.37 & \cellcolor[HTML]{EFEFEF} 14.59 & \cellcolor[HTML]{EFEFEF}  17.70 & \cellcolor[HTML]{EFEFEF} 1408/(1666)\\\hline
FocalClick-B3-S2 & 5.79 & 7.72 & 10.64 & 384/(1666) \\\cline{1-5}
+GAT\cellcolor[HTML]{EFEFEF} & \cellcolor[HTML]{EFEFEF} 5.39 & \cellcolor[HTML]{EFEFEF} 7.34 & \cellcolor[HTML]{EFEFEF} 10.54 & \cellcolor[HTML]{EFEFEF} 311/(1666) \\\hline
SimpleClick-ViT-B & 5.91 & 7.65 & 10.44 & 302/(1666)\\\cline{1-5}
+GAT\cellcolor[HTML]{EFEFEF} &\cellcolor[HTML]{EFEFEF}5.53& \cellcolor[HTML]{EFEFEF}7.05 &\cellcolor[HTML]{EFEFEF}9.57& \cellcolor[HTML]{EFEFEF}251/(1666)\\\hline
\multicolumn{5}{c}{Rice \cite{rice}} \\
 & NoC 80 & NoC 85 & NoC 90 & NoF 90 \\ \hline
 HQ-SAM-ViT-B & 15.68 & 17.50 & 18.95  & 537/(585)\\\cline{1-5}
+GAT\cellcolor[HTML]{EFEFEF} & \cellcolor[HTML]{EFEFEF}15.24 & \cellcolor[HTML]{EFEFEF} 17.25 & \cellcolor[HTML]{EFEFEF}  18.72 & \cellcolor[HTML]{EFEFEF} 535/(585)\\\hline
FocalClick-B3-S2 & 8.62 & 11.78 & 15.90 & 426/(585) \\\cline{1-5}
+GAT\cellcolor[HTML]{EFEFEF} & \cellcolor[HTML]{EFEFEF} 7.96 & \cellcolor[HTML]{EFEFEF} 11.06 & \cellcolor[HTML]{EFEFEF} 15.27 & \cellcolor[HTML]{EFEFEF} 349/(585) \\\hline
SimpleClick-ViT-B & 9.62 & 12.42 & 15.91 & 407/(585)\\\cline{1-5}
+GAT\cellcolor[HTML]{EFEFEF} &\cellcolor[HTML]{EFEFEF}8.74& \cellcolor[HTML]{EFEFEF}11.50 &\cellcolor[HTML]{EFEFEF}15.30& \cellcolor[HTML]{EFEFEF}332/(585)\\\hline
\end{tabular}
}
\end{table}

Experimental results indicate that the proposed algorithm can improve the baseline in most cases.
This experimental result demonstrates that the proposed algorithm also exhibits a significant performance improvement when tested on cross-domain data, further validating its effectiveness as a general click-point enhancement module.

\section{Conclusion}

Our paper introduces a vision transformer-based structured tokens interaction algorithm to overcome weak click control challenges in interactive segmentation.
This paper first utilizes similarity calculation to filter out image tokens that users may focus on, and further employs a GNN model to perform structured aggregation on these specific location tokens, thereby obtaining structured user intent features. Further, dual cross-attention is utilized to fuse the base tokens with the user intent features, thereby achieving structured click control.
To validate the effectiveness of the proposed algorithm, we conducted experiments on three mainstream visual transformer-based algorithms.
Extensive experiments validate the effectiveness and advancement of the proposed algorithm in addressing weak click control issue.

\bibliographystyle{IEEEtran}
\bibliography{refs.bib}

\end{document}